%% file: main.tex
\documentclass[10pt,twocolumn,letterpaper]{article}

\usepackage{cvpr}      
\input{preamble}
\definecolor{cvprblue}{rgb}{0.21,0.49,0.74}
\usepackage[pagebackref,breaklinks,colorlinks,allcolors=cvprblue]{hyperref}
\usepackage{multirow} 
\usepackage{graphicx}
\usepackage{subcaption}
\usepackage{makecell}

\def\confName{CVPR}
\def\confYear{2026}

\title{Reflective Preference Optimization (RPO): Enhancing On-Policy Alignment via Hint-Guided Reflection}

\author{Zihui Zhao\\
Tsinghua Shenzhen International Graduate School, Tsinghua University\\
Shenzhen, China\\
{\tt\small zzh23@mails.tsinghua.edu.cn}
\and
Zechang Li\\
Alibaba Group\\
Hangzhou, Zhejiang\\
{\tt\small lizechang1@huawei.com}
}

\begin{document}
\maketitle
\input{sec/0_abstract}    
\input{sec/1_intro}
\input{sec/2_relatedworks}

\input{sec/3_method}
\input{sec/4_experiments}

\input{sec/5_conclusion}

{
    \small
    \bibliographystyle{ieeenat_fullname}
    \bibliography{ref}
}


\end{document}

%% file: sec/0_abstract.tex
\begin{abstract}
Direct Preference Optimization (DPO) has emerged as a lightweight and effective alternative to Reinforcement Learning from Human Feedback (RLHF) and Reinforcement Learning with AI Feedback (RLAIF) for aligning large language and vision-language models. However, the standard DPO formulation, in which both the chosen and rejected responses are generated by the same policy, suffers from a weak learning signal because the two responses often share similar errors and exhibit small Kullback–Leibler (KL) divergence. This leads to slow and unstable convergence. To address this limitation, we introduce Reflective Preference Optimization (RPO), a new framework that incorporates hint-guided reflection into the DPO paradigm. RPO uses external models to identify hallucination sources and generate concise reflective hints, enabling the construction of on-policy preference pairs with stronger contrastiveness and clearer preference signals. We theoretically show that conditioning on hints increases the expected preference margin through mutual information and improves sample efficiency while remaining within the policy’s distribution family. Empirically, RPO achieves superior alignment with fewer training samples and iterations, substantially reducing hallucination rates and delivering state-of-the-art performance across multimodal benchmarks.
\end{abstract}

%% file: sec/1_intro.tex
\section{Introduction}
\label{sec:intro}

Recent advancements in Large Vision-Language Models (LVLMs) have demonstrated significant capabilities by integrating pre-trained vision encoders with large language models (LLMs).~\cite{alayrac2022flamingo, dai2023instructblip,li2023blip,openai20234v,liu2023visual,liu2024improved,bai2023qwen} These models exhibit strong performance in visual understanding with applications in downstream tasks such as image captioning~\cite{liu2023visual} and medical diagnosis~\cite{hyland2023maira,bannur2024maira}. Nevertheless, their broader practical deployment remains hindered by the issue of \emph{hallucination},~\cite{bai2024hallucination,lan2024survey,liu2024survey} which is prevalent even in the most advanced LVLMs. 

To address this issue, numerous studies employ Reinforcement Learning from Human Feedback (RLHF)~\cite{yu2024rlhf, sun2024aligning} or Reinforcement Learning with AI Feedback (RLAIF)~\cite{zhao2023beyond, xiao2025detecting, yu2025rlaif} to align model responses toward preferred directions. The most widely adopted methods can be categorized into Proximal Policy Optimization (PPO)~\cite{schulman2017proximal} and Direct Preference Optimization (DPO)~\cite{rafailov2023direct}. Although both share the same objective of maximizing rewards derived from the Bradley–Terry model~\cite{bradley1952rank} under the constraint of Kullback–Leibler (KL) divergence~\cite{kullback1951information} between the updated policy and a reference policy, their data usage differs. PPO relies on online rollout data, whereas DPO depends solely on pre-collected offline data~\cite{li2024vlfeedback, wang2024mdpo}. Due to its simplicity and ease of application, DPO is widely used in practice. 
However, the training efficiency of alignment optimization based on DPO-generated preference pairs and the significance of the preference signal within these pairs have been largely overlooked, especially in the context of LVLMs.

For DPO pair construction, current approaches can be categorized into hallucination injection~\cite{zhou2024aligning,sarkar2024mitigating}, hallucination recognition~\cite{feng2024mixture,zhao2023beyond, xiao2025detecting}, and self-evolution~\cite{yu2025rlaif}. Among them, the self-evolution paradigm often outperforms the others because it exclusively uses on-policy preference pairs, whereas the other two involve either rejected or preferred responses generated off-policy. However, the self-evolution paradigm also faces a fundamental limitation that both the \textit{chosen} and \textit{rejected} responses are typically generated by the same policy~\cite{yu2025rlaif}, often sharing stylistic and factual biases. The KL divergence between them is small and the preference margin weak, leading to slow convergence and limited preference signals. 

This limitation is particularly evident in multimodal settings, where LVLMs frequently exhibit \emph{hallucinations} and produce semantically fluent but factually inaccurate image-grounded descriptions. Recent work has shown that injecting external corrections can greatly enhance preference learning~\cite{feng2024mixture,zhao2023beyond, xiao2025detecting}. While this approach yields notable improvements in hallucination mitigation, it also relies on human-crafted or model-edited responses that lie partially outside the policy’s support. This off-policy component introduces additional complexity and weakens the theoretical guarantees of DPO’s on-policy optimization. 

\begin{figure}
    \centering
    \includegraphics[width=1\linewidth]{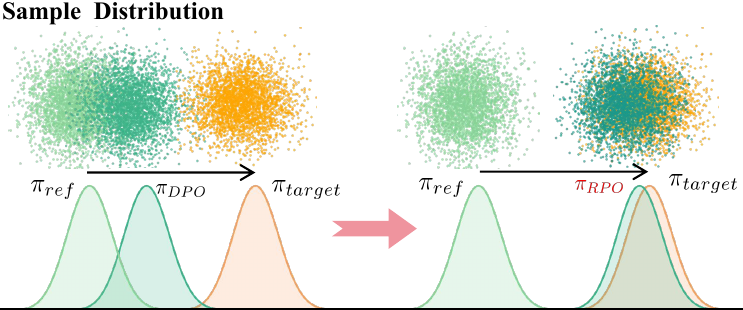}
    \caption{Illustration of the sample and policy distributions for standard DPO and the proposed RPO under the same preference optimization direction. RPO produces samples that align more closely with the target policy, demonstrating stronger and more effective preference optimization.}
    \label{fig:d}
\end{figure}

In this work, we aim to enhance the preference signal of DPO-based pairs while strictly remaining within the on-policy learning regime. We introduce \textbf{Reflective Preference Optimization (RPO)}, a novel paradigm that enables models to perform self-reflective preference learning. As illustrated in Fig.~\ref{fig:d}, RPO systematically increases the preference margin while preserving on-policy consistency. RPO incorporates a hint-guided reflection mechanism that strengthens the discriminability of preference pairs, maintains full on-policy consistency, and operates without human assistance.


Concretely, after the policy $\pi_\theta$ generates an initial response $y^-$ for an input $x$, an external critique model $C$ (such as GPT-4V or GLM-4.5V) analyzes the pair $(x, y^-)$ and produces a focused \textit{hint} $h=C_\phi(x,y^-)$ identifying the likely error or suggesting a correction direction. The same policy then generates a new response $y^+$ conditioned on $(x, h)$. This second output, guided by reflection, typically corrects the factual or logical mistakes present in $y^-$ while retaining the stylistic characteristics of $\pi_\theta$. The pair $(y^+, y^-)$ thus forms an \emph{on-policy preference pair} that preserves the model’s distributional coherence but exhibits a significantly larger preference margin. Therefore, the method remains entirely on-policy, avoiding distributional collapse or unstable KL terms that arise in off-policy corrections.

Extensive experiments on standard LVLM hallucination benchmarks show that our proposed RPO consistently surpasses existing methods, even when trained on smaller datasets. In summary, this work makes the following contributions:
\begin{itemize}
    \item We introduce \textbf{Reflective Preference Optimization (RPO)}, a simple yet effective extension of DPO that leverages hint-guided self-reflection to enhance preference separability while remaining on-policy.
    \item We provide a theoretical framework that connects reflection-induced mutual information with preference margin amplification and establishes formal conditions under which this amplification occurs.
    
    \item We empirically validate RPO across hallucination benchmarks, achieving state-of-the-art hallucination mitigation.
\end{itemize}

%% file: sec/2_relatedworks.tex
\section{Related Work}
\label{sec:related}

\paragraph{Preference Alignment.}
As a key technique driving recent advances in large language and vision–language models (LLMs/LVLMs), \textit{Reinforcement Learning from Human Feedback} (RLHF)~\cite{christiano2017deep, ouyang2022training} has proven highly effective for aligning fine-tuned models with human preferences. By leveraging large-scale human or AI preference data and reinforcement learning algorithms, RLHF has enabled the success of major model families such as GPT\cite{brown2020language,achiam2023gpt}, LLaMA~\cite{touvron2023llama}, Qwen~\cite{bai2023qwen}, Gemini~\cite{team2024gemini}, and Claude~\cite{TheC3}. The original RLHF formulation relies on Proximal Policy Optimization (PPO)\cite{schulman2017proximal}, which provides stability but depends heavily on a reward model and numerous hyperparameters. This complexity has motivated the exploration of simpler, reward-free alternatives. \textit{Direct Preference Optimization} (DPO)~\cite{rafailov2023direct} emerged as an appealing surrogate that removes explicit reward modeling while maintaining competitive alignment quality. However, DPO still trails PPO in certain benchmarks~\cite{xu2024contrastive}, prompting subsequent variants such as ORPO~\cite{hong2024orpo}, TPO~\cite{li2025test}, and SimPO~\cite{meng2024simpo}, which relax reference constraints or modify objectives to improve stability and generalization. More relevant to our work, iterative and self-play variants~\cite{yu2024rlaif} attempt to resolve off-policy issues by sampling preference pairs on-policy, an idea also adopted by RLAIF-V~\cite{yu2024rlhf}. Nonetheless, these approaches often struggle to efficiently mitigate persistent hallucinations in multimodal contexts.

\paragraph{DPO and Variants.}
Building on preference alignment, DPO~\cite{rafailov2023direct} formulates the alignment objective as a Bradley–Terry likelihood ratio between preferred and rejected responses, implicitly regularized by a KL term to the reference policy. It removes the need for explicit reward modeling and online rollouts, enabling scalable alignment for both text and vision–language domains.
Recent efforts on pair construction can be divided into: (1) Injection-based methods generate negatives by corrupting clean answers~\cite{zhou2024aligning,sarkar2024mitigating}; (2) Recognition-based approaches rely on stronger models or humans to correct outputs~\cite{feng2024mixture,zhao2023beyond, xiao2025detecting}; (3)Self-evolution strategies sample both responses from the same policy for labeling~\cite{yu2025rlaif}. Our proposed \textbf{Reflective Preference Optimization (RPO)} follows the self-evolution path but introduces a \emph{hint-guided reflection} mechanism that regenerates the preferred response under explicit critique, yielding higher contrast without departing from on-policy consistency. Unlike recognition-based methods, RPO does not modify responses off-policy; the critique model provides only a concise hint, while the reflective response is still generated by the same policy.

\paragraph{Hallucination in LVLMs.}
Hallucination—text inconsistent with visual or contextual evidence—remains a major challenge for LVLMs. RL-free methods address hallucination through architecture or decoding adjustments, such as visual contrastive decoding, token-level penalties, and attention diagnostics~\cite{manevich2024mitigating,tang2025mitigating,liu2025multi,han2024skip}. RL(HF/AIF)-based methods instead learn to avoid hallucination directly from preference supervision. Due to its simplicity and scalability, DPO has become the dominant framework for hallucination mitigation in LVLMs~\cite{xiao2025detecting,sun2024aligning,wang2024mdpo,yu2025rlaif}. Our RPO approach belongs to this family but explicitly increases the conditional mutual information between the reflective hint and the preferred response, producing stronger preference margins and reducing hallucination without deviating from the model’s own distribution. This provides a principled and fully on-policy approach to preference alignment, which has been largely missing in prior LVLM hallucination literature.

%% file: sec/3_method.tex
\section{Method}

In this section, we first revisits the DPO formulation and the limitations of the self-evolution paradigm. Then we introduce the proposed Reflective Preference Optimization (RPO) framework together with its theoretical interpretation, followed by the data construction pipeline and optimization objectives.

\subsection{Preliminaries and Limitations of Self-Evolution}

Given an input $x$ and a policy $\pi_\theta$, DPO~\cite{rafailov2023direct} learns to prefer a better response $y^+$ over a worse one $y^-$ by maximizing a logistic likelihood ratio:
\begin{equation}
\begin{aligned}
\mathcal{L}_{\text{DPO}}
=& -\mathbb{E}_{(x,y^+,y^-)\sim \mathcal{D}}\\
&\Big[
\log\sigma\!\Big(
\beta \!\Big(
\log\frac{\pi_\theta(y^+|x)}{\pi_{\text{ref}}(y^+|x)} 
-\log\frac{\pi_\theta(y^-|x)}{\pi_{\text{ref}}(y^-|x)}
\Big)
\Big)
\Big]
\end{aligned}
\end{equation}
where $\pi_{\text{ref}}$ is a frozen reference model and $\beta$ controls the effective temperature of the implicit KL regularization.
This objective aligns the model by enlarging the preference margin
\begin{equation}
\Delta\ell(x)=
\log\pi_\theta(y^+|x)-\log\pi_\theta(y^-|x)
\end{equation}
Recent work~\cite{yang2025mitigating} demonstrates that keeping both samples on-policy is essential for stable optimization. However, the \emph{self-evolution} paradigm, where $y^+$ and $y^-$ are drawn from the same policy, introduces a key limitation that the generated responses are often stylistically and semantically similar. As a result, most pairs exhibit a small expected margin $\mathbb{E}[\Delta\ell(x)]$, contributing limited gradient signal. Formally, the model’s log-probability can be written as
\begin{equation}
\log\pi_\theta(y|x)=s_\theta(x,y)+\epsilon
\end{equation}
where $s_\theta(x,y)$ denotes the deterministic score and $\epsilon\!\sim\!\mathcal{N}(0,\sigma^2)$ models stochastic decoding variation. For a self-evolution based preference pair $(y^+,y^-)$ sampled from $\pi_\theta$, the expected margin satisfies
\begin{equation}
\mathbb{E}[\Delta\ell(x)] \!\to\! 0,\quad
\mathrm{Var}[\Delta\ell(x)] \!\to\! 2\sigma^2
\end{equation}
Thus, while self-evolution ensures on-policy sampling, it tends to yield near-zero mean margins and high-variance gradients, producing weak preference and optimization signals. Empirically, such pairs exhibit low inter-sample divergence:
\begin{equation}
\mathrm{KL}\!\big(\pi_\theta(\cdot|x,y^+)\,\|\,\pi_\theta(\cdot|x,y^-)\big)\!\ll\!1
\end{equation}
which indicates that the policy distributions over preferred and rejected responses are largely overlapped. This low-contrast regime leads to a flat optimization landscape and slow convergence.

\subsection{Reflective Preference Optimization Framework}

\begin{figure*}
    \centering
    \includegraphics[width=1\linewidth]{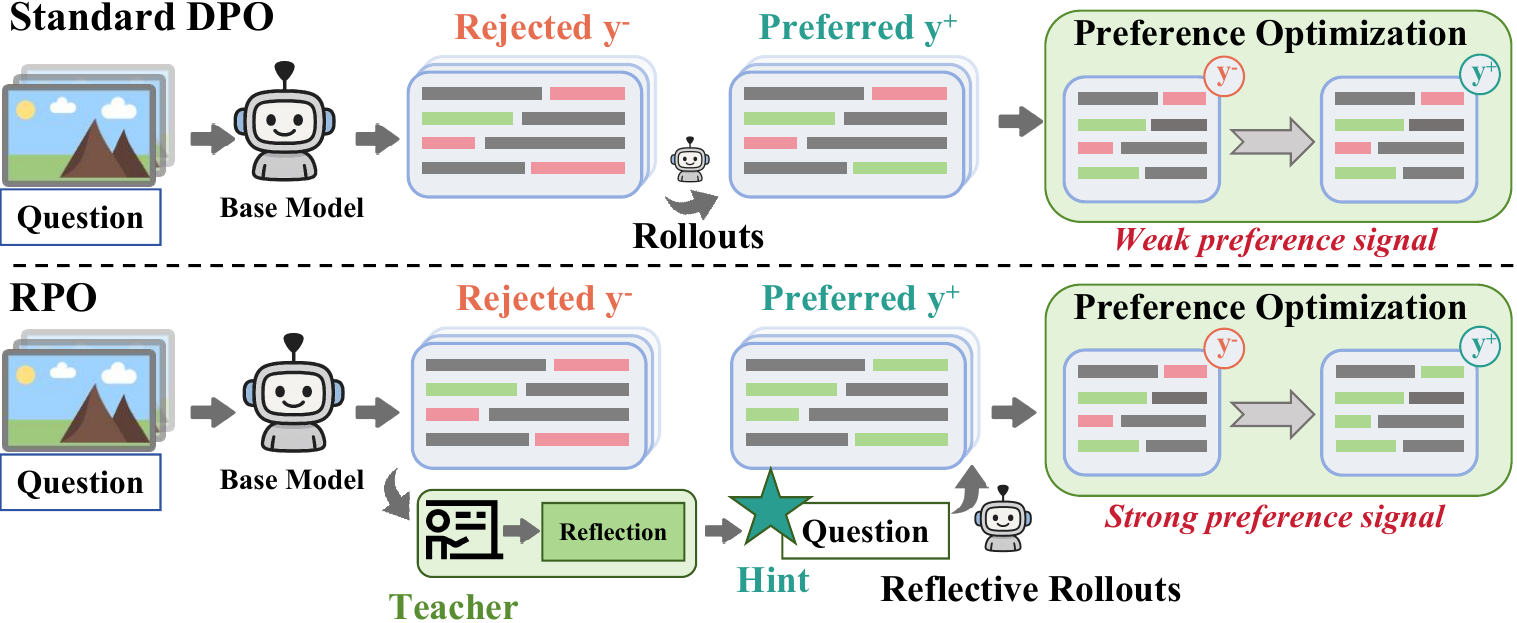}
    \caption{Illustration of the proposed RPO framework. Compare to standard DPO, which generates both preferred and rejected responses from identical query inputs, RPO generates preferred responses conditioned on reflective hints provided by the teacher model. This design yields a stronger preference signal and facilitates more effective preference optimization.}
    \label{fig:rpo method}
\end{figure*}
To mitigate this limitation, as shown in the Figure.~\ref{fig:rpo method}, RPO introduces a reflection–based perturbation mechanism that increases contrast between preference pairs while remaining strictly on-policy paradigm.  

Given an input $x$, the model first generates a base response $y^-\!\sim\!\pi_\theta(\cdot|x)$. An external critique model $C_\phi$ then analyzes $(x,y^-)$ and produces a compact hint $h=C_\phi(x,y^-)$ that summarizes the main error or correction direction (e.g., ''focus on the object color in the image''). Conditioned on the reflective hint, the same policy regenerates
\begin{equation}
y^+\sim \pi_\theta(\cdot|x,h)
\end{equation}

This process introduces a conditional perturbation that shifts the policy toward a more informative region of its distribution. Under a small-perturbation assumption, the reflected log-probability can be approximated as
\begin{equation}
\log\pi_\theta(y^+|x,h)
\simeq
\log\pi_\theta(y^+|x)
+\delta_\theta(x,h)
\end{equation}
where $\delta_\theta(x,h)$ denotes the reflection-induced bias term. This approximation holds under mild smoothness assumptions on the policy with respect to input prompts. The expected preference margin becomes
\begin{equation}
\begin{aligned}
\mathbb{E}[\Delta\ell_h(x)]
&= \mathbb{E}[\log\pi_\theta(y^+|x,h)-\log\pi_\theta(y^-|x)] \\
&= \mu_h\\ \quad \mu_h &= \mathbb{E}[\delta_\theta(x,h)]>0
\end{aligned}
\end{equation}

Hence, reflection systematically shifts the expected margin toward positive values, improving both the magnitude and stability of the learning signal without breaking on-policy constraints.

From an optimization perspective, the reflective hint $h$ perturbs the local likelihood landscape of $\pi_\theta$ while staying within the same policy manifold.  
For a pair $(y^+,y^-)$, the original DPO gradient is
\begin{equation}
g(x,y^+,y^-)=\nabla_\theta\log\frac{\pi_\theta(y^+|x)}{\pi_\theta(y^-|x)}
\end{equation}
while the RPO gradient becomes
\begin{equation}
g_h(x,y^+,y^-)=\nabla_\theta\log\frac{\pi_\theta(y^+|x,h)}{\pi_\theta(y^-|x)}
\end{equation}
Since reflection affects only the conditioning context, both responses are drawn from the same policy, ensuring on-policy learning. This introduces an additional structured variance term:
\begin{equation}
\mathrm{Cov}_h[g_h]
\simeq
\mathrm{Cov}[g] + \Sigma_h
\end{equation}
where $\Sigma_h$ captures the curvature induced by the reflective hint. As $\mathbb{E}[\delta_\theta(x,h)]>0$, this perturbation amplifies informative gradient components and accelerates convergence.  
The induced divergence between reflected and base distributions is
\begin{equation}
\mathrm{KL}\!\big(\pi_\theta(\cdot|x,h)\,\|\,\pi_\theta(\cdot|x)\big)
=
\mathbb{E}_{y\sim\pi_\theta(\cdot|x,h)}[\delta_\theta(x,h)]
\end{equation}
which functions as an adaptive margin regularizer—expanding the effective preference gap while keeping optimization well-conditioned.  
Because both responses originate from $\pi_\theta$, gradient updates remain consistent and non-degenerate throughout training. The hint acts as a controlled, on-manifold perturbation that enhances contrast, stabilizes optimization, and improves sample efficiency.

\subsection{Data Collection and On-Policy Alignment}

\begin{figure}
    \centering
    \includegraphics[width=1\linewidth]{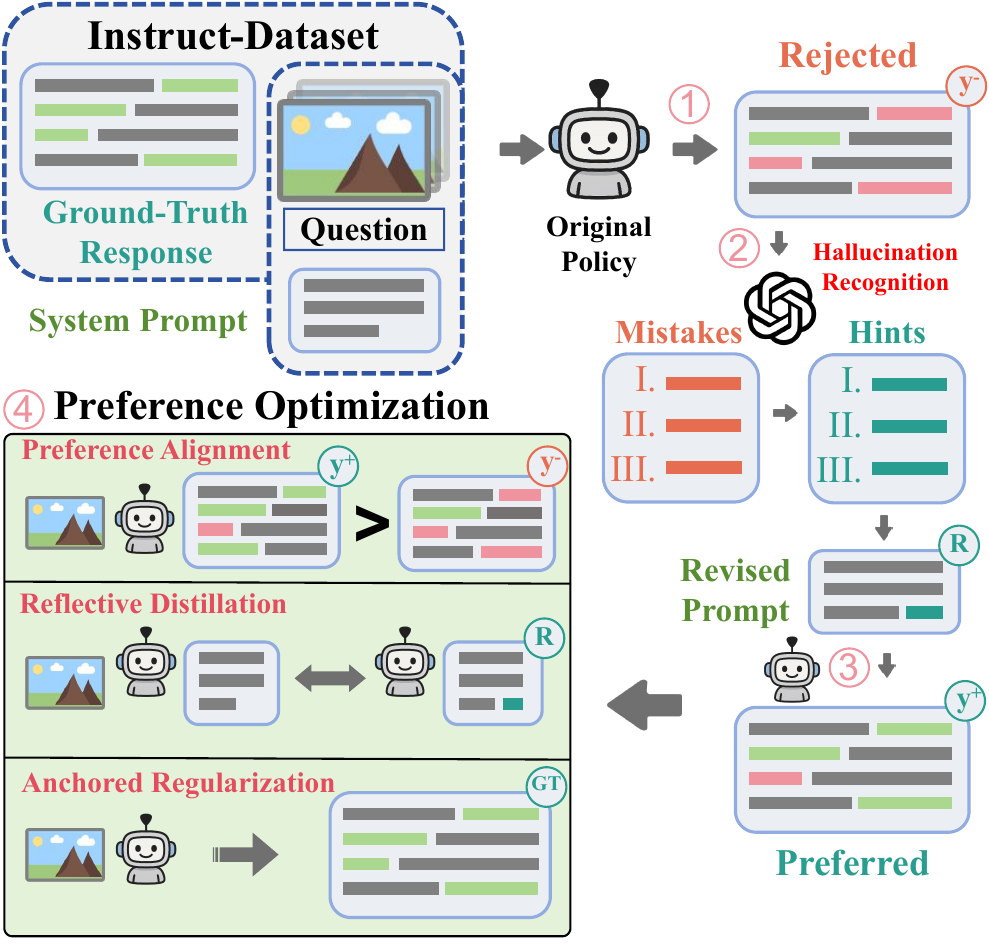}
    \caption{The proposed RPO framework consists of four steps: (1) generating responses from the original policy using image–prompt inputs; (2) identifying hallucinations and producing reflective hints using a teacher model; (3) appending the hint to the system prompt and regenerating the preferred response with the same policy; (4) performing preference optimization on the current policy using the constructed preference pairs.}
    \label{fig:method_detail}
\end{figure}

The success of RPO depends on constructing preference pairs that are both contrastive and on-policy. The general pipeline of our RPO method is shown in Figure.~\ref{fig:method_detail}.

Given an image–text input $x=(x_m, x_t)$, the model first generates a response $y_{\text{Gen}}\sim\pi_\theta(\cdot \mid x_m, x_t)$. An external model $C_\phi$ (e.g., GPT-4V or GLM-4.5V) compares the generated response $y_{\text{Gen}}$ with the ground-truth response $y_{\text{GT}}$, identifies hallucinated segments $y^-$, and produces a reflective hint $h = C_\phi(x, y^-)$ that based on the detected mistakes. Unlike a full rewriting, the hint is concise and localized, serving as a first-order correction within the prompt space while preserving the model’s generative distribution. It is also crucial that the hint does not impose any hard supervision on token sequences. In addition, $C_\phi$ assigns each hallucinated segment a severity weight $W_{\text{hal}}$, indicating the magnitude of the hallucination associated with that segment. 

Conditioned on $(x, h)$, the policy then regenerates a reflective response $y^+\sim\pi_\theta(\cdot \mid x, h)$. Since both responses are sampled from the same policy, RPO strictly preserves the on-policy alignment property. It is worth noting that using the ground-truth response $y_{\text{GT}}$ as the preferred sample is generally impractical, because $y_{GT} \not\sim \pi_\theta$, the term $\log \frac{\pi_\theta(y^+)}{\pi_{\text{ref}}(y^+)}$ collapses toward large negative values, causing DPO gradients to degenerate.
The resulting dataset
\begin{equation}
\mathcal{D}_{\text{RPO}} = \{(x,y^+,y^-,h)\}
\end{equation}
forms an on-policy reflective corpus that captures both the model’s self-generated weaknesses ($y^-$) and its hint–guided corrections ($y^+$), enabling contrastive yet stable preference optimization.

\subsection{RPO Training}

With the reflective corpus established, RPO is trained under a unified objective combining Preference Alignment, Reflective-Distillation, and Anchored Regularization. The goal is to encourage reflective responses while transferring improvements back to the base policy.

\textbf{Preference Alignment.}
The primary objective follows the DPO formulation, weighted by hallucination severity $W_{\text{hal}}$:
\begin{equation}
\begin{aligned}
\mathcal{L}_{\text{Pref}}
=&
-\mathbb{E}_{(x,y^+,y^-,h)\sim \mathcal{D}_{\text{RPO}}}\\
&\log\sigma\!\Big(
\beta \big[
W_{\text{hal}}\log\frac{\pi_\theta(y^+|x)}{\pi_{\text{ref}}(y^+|x)}
-
\log\frac{\pi_\theta(y^-|x)}{\pi_{\text{ref}}(y^-|x)}
\big]
\Big)
\end{aligned}
\end{equation}

\textbf{Reflective-Distillation.}
To transfer improvements from reflective to unconditioned generation, we introduce
\begin{equation}
\mathcal{L}_{\text{RD}}
= \mathbb{E}_{x,h}
D_{\text{KL}}\big(\pi_\theta(\cdot|x,h)\,\|\,\pi_\theta(\cdot|x)\big)
\end{equation}

This encourages the model to internalize the improved reasoning without relying on hints at inference time.

\textbf{Anchored Regularization.}
Following mDPO~\cite{wang2024mdpo}, we add an anchoring term to prevent degradation of preferred likelihoods:
\begin{equation}
\mathcal{L}_{\text{Anc}}
=
-\mathbb{E}_{(x,y_{\text{GT}})\sim \mathcal{D}_{\text{RPO}}}
\log\sigma\!\Big(
\beta 
\log\frac{\pi_\theta(y_{\text{GT}}|x)}{\pi_{\text{ref}}(y_{\text{GT}}|x)} -\delta
\Big)
\end{equation}

The final objective combines these terms:
\begin{equation}
\mathcal{L}_{\text{RPO}} =
\mathcal{L}_{\text{Pref}} + \lambda_1 \mathcal{L}_{\text{RD}} + \lambda_2 \mathcal{L}_{\text{Anc}}
\end{equation}

This joint training enables reflection to improve both conditional and base reasoning. The distilled objective allows the policy to internalize reflective corrections, leading to improved grounding and reasoning even without hints at inference time. As a result, RPO provides a scalable and self-improving preference optimization framework that achieves superior performance with fewer samples and no manual post-processing.

%% file: sec/4_experiments.tex
\section{Experiments}

\subsection{Experiment Setup}

\textbf{Models and Datasets.} We apply RPO to two LVLMs of different scales: LLaVA-v1.5-7B and LLaVA-v1.5-13B~\cite{liu2023llava}, both equipped with CLIP ViT-L/336px~\cite{radford2021learning} as the vision encoder. The 7B and 13B variants are based on Vicuna-7B and Vicuna-13B, respectively. Each model was originally pretrained on 558K image–text pairs and subsequently fine-tuned on 665K instruction-following samples.
For RPO training, we randomly sample 4K instances from the RLAIF-V dataset~\cite{yu2025rlaif} and use their preferred responses as surrogate ground truth.

\textbf{Evaluation Metrics.} We focus on hallucination mitigation in LVLMs and conduct evaluations on four widely used benchmarks: 1. \textbf{AMBER}~\cite{wang2023amber}: A generative benchmark with 1004 images and fine-grained object annotations. Using the official codebase, we report CHAIR score, object coverage, hallucination rate, and cognitive alignment; 2. \textbf{MMHalBench}~\cite{sun2024aligning}: A QA benchmark with 96 images across 12 categories. Following the official protocol, GPT-4 rates responses from 0 to 6, and hallucination rate is computed as the proportion of samples scoring below; 3. \textbf{Object HalBench}~\cite{rohrbach2018object}: A standard benchmark for object hallucination. We evaluate 300 test instances using the implementation from~\cite{yu2024rlhf}, reporting hallucination rates at the response level (CHAIRs) and object level (CHAIRi); 4. \textbf{POPE}~\cite{li2023evaluating}: A yes/no QA benchmark for object hallucination. We report accuracy and precision on its 3000-instance Adversarial split.

\textbf{Baselines.} We compare RPO against representative RLHF/RLAIF- and DPO-based methods, including HALVA~\cite{sarkar2024mitigating}, POVID~\cite{zhou2024aligning}, RLHF-V~\cite{yu2024rlhf}, HA-DPO~\cite{zhao2023beyond}, HSA-DPO~\cite{xiao2025detecting}, RLAIF-V~\cite{yu2025rlaif}, mDPO~\cite{wang2024mdpo}, and OPA DPO~\cite{yang2025mitigating}.

\textbf{Implementation Details.} For both 7B and 13B models, we train RPO for 4 epochs with batch size 4, using LoRA with rank 256 and scaling factor 512. For the anchored preference objective, we set $\beta = 0.1$ and $\delta = 0$. For the combined loss, we use $\gamma_1 = \gamma_2 = 0.2$.

\subsection{Preference-Pair Evaluation}

\begin{figure}
    \centering
    \includegraphics[width=0.9\linewidth]{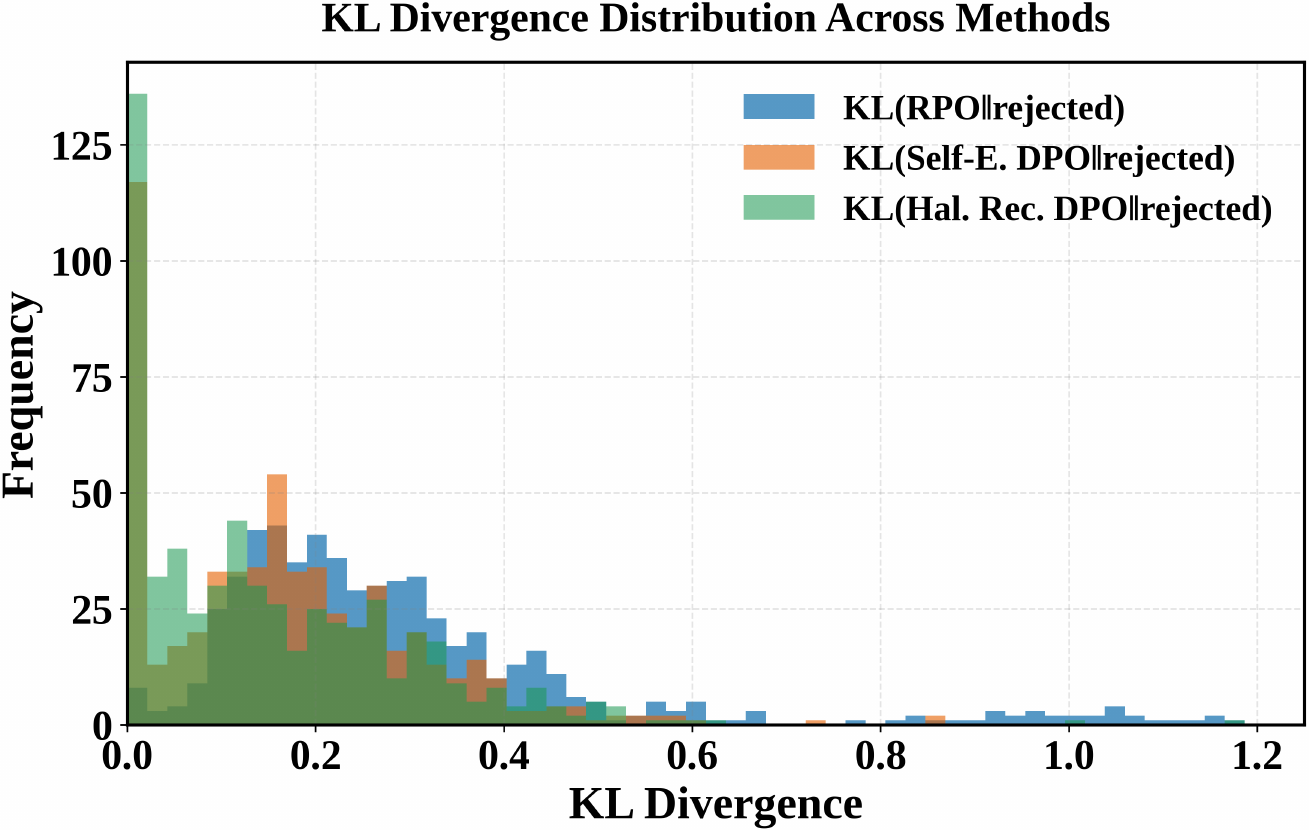}
    \caption{KL divergence distributions of preference pairs constructed by RPO, Self-Evolution DPO (Self-E. DPO), and Hallucination-Recognition DPO (Hal. Rec. DPO), all based on the same rejected responses. The blue bars (RPO) show consistently higher divergence and contrastiveness, indicating that RPO produces preference pairs with stronger preference signals and more discriminative alignment behavior.}
    \label{fig:kl}
\end{figure}

To assess the effectiveness of reflection-guided preference pair construction, we compute the KL divergence between preferred and rejected responses, $\mathrm{KL}(x_{\text{preferred}} \;\|\; x_{\text{rejected}})$, and compare three paradigms under identical inputs and identical rejected responses: RPO, self-evolution DPO, and hallucination-recognition DPO. As shown in Figure~\ref{fig:kl}, RPO generates preference pairs with substantially higher KL divergence than the other two approaches, indicating stronger preference contrast and a more decisive alignment signal. The numerical statistics in Table~\ref{tab:kl} further confirm that RPO consistently achieves higher mean, median, and percentile KL values, demonstrating that reflection produces preference pairs with clearer, more informative supervisory signals that facilitate more stable and efficient optimization.

\begin{table}[t]
\centering
\caption{Comparison of the average, median, 25th percentile, and 75th percentile KL-divergence values across different preference pair construction methods, all evaluated on the same set of rejected responses. RPO consistently achieves higher divergence statistics, indicating stronger preference contrast and better alignment signal quality.}
\label{tab:comparison}
\begin{tabular}{lcccc}
\toprule
\textbf{$\mathrm{KL}$} & \textbf{Mean} & \textbf{Median} & \textbf{P25} & \textbf{P75}\\
\midrule
RPO & 0.293 & 0.237 & 0.161 &0.354 \\
Self-E. & 0.172 & 0.158 & 0.061 &0.251 \\
Hal. Rec. & 0.153 & 0.118 & 0.026 &0.238 \\
\bottomrule
\label{tab:kl}
\end{tabular}
\end{table}

\subsection{Policy Distribution}

\begin{figure}
    \centering
    \includegraphics[width=1\linewidth]{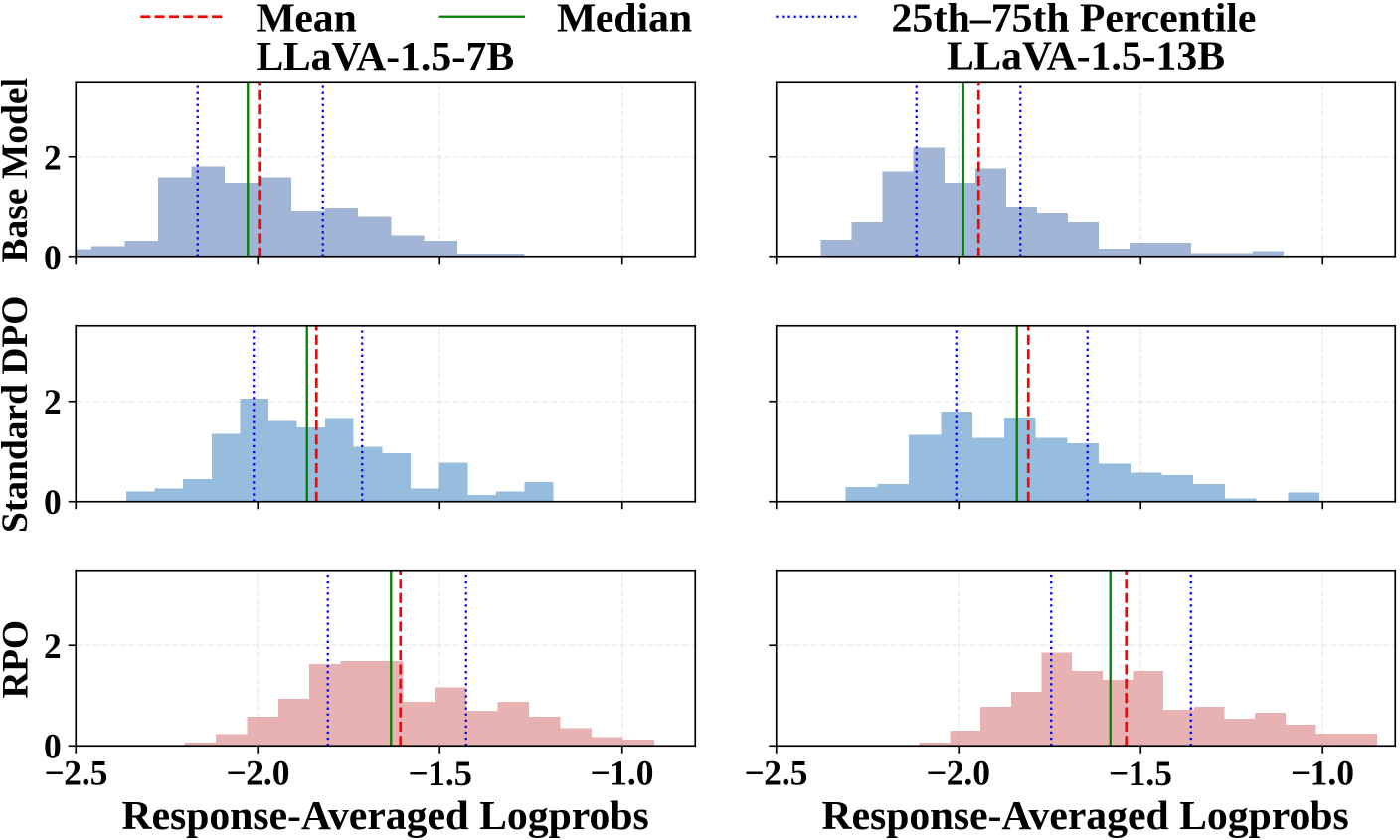}
    \caption{Distribution of response-averaged log probabilities for standard (ground-truth) responses across different models. The RPO model exhibits a clear right-shifted distribution, indicating a significantly higher likelihood of generating the correct responses compared with other methods.}
    \label{fig:logporbs}
\end{figure}

To validate that standard DPO fails to achieve effective preference alignment while our proposed RPO framework resolves this limitation, we visualize the response-averaged token log-probabilities of the ground-truth (standard) responses, computed as $\frac{1}{L} \sum_{i=1}^{L} \log \pi(y_i \mid x)$. As illustrated in the Figure~\ref{fig:logporbs}, the distribution shift induced by standard DPO remains marginal, indicating limited preference alignment. In contrast, the RPO-trained model exhibits a pronounced rightward shift in the distribution, revealing a substantial enhancement in aligning model preferences toward the ground-truth responses.

\begin{table*}[h]
\centering
\caption{Comparison of RLAIF- and RLHF-based algorithms for enhancing large vision-language models (LVLMs). All evaluations adopt greedy sampling for inference. Results of baseline methods are reported from their respective original papers. The best performance within each group is highlighted in \textbf{bold}.}
\resizebox{\textwidth}{!}{
\begin{tabular}{l c cccc cc cc cc}
\toprule
\multirow{2}{*}{\textbf{Algorithm}} & \multirow{2}{*}{\textbf{Feedback}} &
\multicolumn{4}{c}{\textbf{AMBER (1004)}} &
\multicolumn{2}{c}{\textbf{MMHal-Bench (96)}} &
\multicolumn{2}{c}{\textbf{Object Hal (300)}} &
\multicolumn{2}{c}{\textbf{POPE Adv. (3000)}} \\
\cmidrule(lr){3-6} \cmidrule(lr){7-8} \cmidrule(lr){9-10} \cmidrule(lr){11-12}
 &  & CHAIR$\downarrow$ & Cover$\uparrow$ & HalRate$\downarrow$ & Cog$\downarrow$
 & Score$\uparrow$ & HalRate$\downarrow$
 & CHAIRs$\downarrow$ & CHAIRl$\downarrow$
 & Acc.$\uparrow$ & Pre.$\uparrow$ \\
\midrule
\midrule
\multicolumn{12}{l}{\textbf{LLaVA-Instruct-1.5-7B [5,6]}} \\
(base) & --                    & 7.7 & 51.6 & 34.7 & 4.2 & 2.01 & 0.61 & 55.67 & 15.96 & \textbf{84.93\%} & 89.10\% \\
+LLaVA-RLHF [21] & Self-Reward & 9.7 & 53.2 & 46.6 & 5.3 & 1.88 & 0.71 & 58.00 & 15.61 & 80.00\% & 87.19\% \\
+HALVA [17]   & GPT-4V      & 6.6 & \textbf{53.0} & 32.2 & 3.4 & 2.25 & 0.54 & 41.40 & 11.70 & -- & -- \\
+mDPO [27] & GPT-4V    & 4.4 & 52.4 & 24.5 & 2.4 & 2.39 & 0.54 & 35.70 & 9.80  & -- & {95.36\%} \\
+HA-DPO [18]      & GPT4        & 7.8 & 52.1 & 35.6 & 4.2 & 1.89 & 0.65 & 54.00 & 14.45 & 84.90\% & 90.42\% \\
+POVID [16]       & GPT-4V      & 7.4 & 51.3 & 34.3 & 3.9 & 2.08 & 0.60 & 50.67 & 15.28 & 84.77\% & 89.01\% \\
+RLAIF-V [20]     & LLaVA-Next  & 3.0 & 50.4 & 16.2 & \textbf{1.0} & {3.00} & \textbf{0.38} & 16.00 & \textbf{3.70} & 81.57\% & 94.97\% \\
{+OPA-DPO}        & GPT-4V      & {2.2} & 47.9 & {11.6} & \textbf{0.9} & 2.83 & 0.45 & {13.00} & 4.25 & 82.60\% & {95.61\%} \\
{+RPO }        & GPT-4V      & \textbf{2.0} & {49.5} & \textbf{10.5} & \textbf{0.9} & \textbf{2.95} & {0.41} & \textbf{12.50} & {3.90} & {83.20\%} & \textbf{95.80\%} \\
\midrule
\multicolumn{12}{l}{\textbf{LLaVA-Instruct-1.5-13B [5,6]}} \\
(base) & --                    & 6.8 & 51.9 & 31.8 & 3.3 & 2.48 & 0.52 & 51.00 & 13.71 & 85.50\% & 90.31\% \\
+LLaVA-RLHF [21] & Self-Reward & 7.7 & 52.3 & 38.6 & 4.0 & 2.27 & 0.64 & 44.67 & 11.83 & 82.47\% & 90.25\% \\
+RLHF-V (HD) [15] & Human     & 6.3 & 46.1 & 25.1 & 2.1 & 2.81 & 0.49 & -- & -- & -- & -- \\
+HSA-DPO [19]  & GPT-4V/4.1 & \textbf{2.1} & 47.3 & 13.4 & 1.2 & 2.61 & 0.48 & -- & -- & 84.00\% & 80.20\% \\
+HALVA [17]    & GPT-4V      & 6.4 & 52.6 & 30.4 & 3.2 & 2.58 & 0.45 & 45.40 & 12.80 & -- & -- \\
{+OPA-DPO}      & GPT-4V      & {2.4} & 48.3 & {12.8} & {0.9} & {3.07} & {0.39} & {16.33} & \textbf{5.48} & 82.63\% & {96.31\%} \\
{+RPO }        & GPT-4V      & {2.2} & \textbf{53.3} & \textbf{11.9} & \textbf{0.8} & \textbf{3.05} & \textbf{0.37} & \textbf{15.50} & \textbf{5.49} & {83.80\%} & \textbf{96.50\%} \\
\bottomrule
\label{tab:main_results}
\end{tabular}
} 
\end{table*}

\subsection{Evaluation Results}
The quantitative results across multiple benchmarks are summarized in Table~\ref{tab:main_results}. Our proposed RPO framework consistently outperforms all baseline methods on both LLaVA-Instruct-1.5-7B and LLaVA-Instruct-1.5-13B models. Notably, RPO achieves superior performance with substantially fewer training epochs, particularly on the 13B variant. This efficiency can be attributed to the high-contrastive preference pairs and stronger preference signals introduced by the RPO paradigm, which promote faster and more stable optimization.

Moreover, the LLaVA-Instruct-1.5-13B model exhibits more pronounced improvements under RPO training, suggesting that larger models are better able to leverage reflective hints to generate preferred responses.

\begin{figure}
    \centering
    \includegraphics[width=1\linewidth]{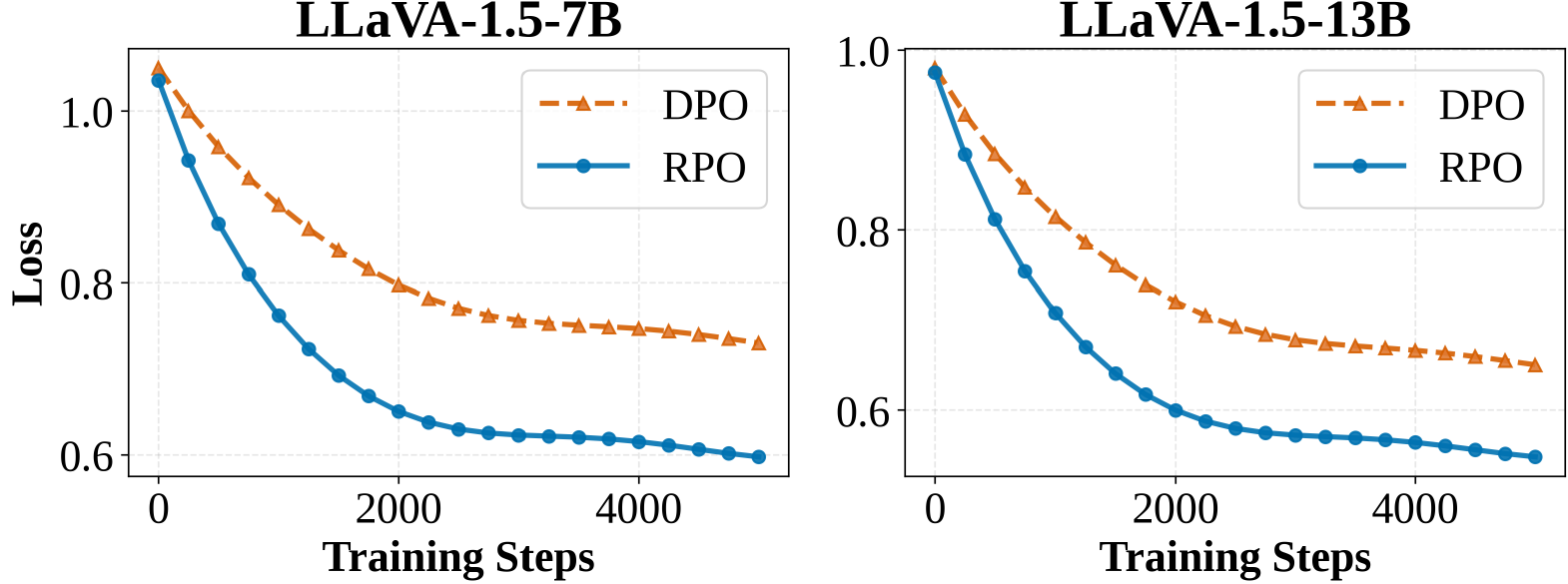}
    \caption{Training loss curves of RPO and standard DPO on LLaVA-Instruct-1.5-7B and LLaVA-Instruct-1.5-13B models. The RPO method demonstrates faster convergence and achieves lower final loss values, indicating more efficient and stable optimization.}
    \label{fig:convergence_curve}
\end{figure}

To further assess training dynamics, we compare the convergence behavior of DPO and RPO during preference optimization. As illustrated in Figure~\ref{fig:convergence_curve}, RPO demonstrates faster convergence and lower final loss values, requiring significantly fewer optimization steps to stabilize. This advantage stems from the high-quality preference pairs constructed under hint guidance.

However, the convergence and final performance remain sensitive to the quality of these hints as accurate and informative hints effectively guide the model toward improved responses, while vague or ambiguous ones may provide limited benefit. To systematically investigate this dependency, we perform ablation studies to analyze model performance under different hint sources and qualities.

\subsection{Ablation Studies}

\begin{table}[t]
\centering
\scriptsize
\setlength{\tabcolsep}{1pt}
\renewcommand{\arraystretch}{1.15}
\begin{tabular}{llcccc|cc}
\toprule
\multicolumn{2}{c}{} & \multicolumn{4}{c}{\textbf{AMBER}} & \multicolumn{2}{c}{\textbf{Object Hal}}\\
\cmidrule(lr){3-6}\cmidrule(l){7-8}
\textbf{\makecell{Model\\size}} & \textbf{Ablation} & \textbf{CHAIR}\,$\downarrow$ & \textbf{Cover}\,$\uparrow$ & \textbf{HalRate}\,$\downarrow$ & \textbf{Cog}\,$\downarrow$ & \textbf{CHAIRs}\,$\downarrow$ & \textbf{CHAIRi}\,$\downarrow$\\
\midrule
\multirow{5}{*}{\textbf{7B}} &  RPO                           & 2.0 & 49.5 & 10.5 & 0.9  & 12.50 & 3.90 \\
                              & w/o $y^+$                      & 4.5 & 44.3 & 16.1 & 1.4  & 15.44 & 9.87 \\
                              & w/o Anc                        & 2.7 & 45.7 & 13.2 & 1.2  & 14.32 & 8.74 \\
                              & w/o RD                         & 2.4 & 47.1 & 12.1 & 1.1  & 13.37 & 6.41 \\
                              & w/o RD\&Anc                    & 2.6 & 45.4 & 13.7 & 1.2  & 14.59 & 9.06 \\
\midrule
\multirow{5}{*}{\textbf{13B}} &  RPO                           & 2.2 & 53.3 & 11.9 & 0.8  & 15.50 & 5.49 \\
                              & w/o $y^+$                      & 4.6 & 45.9 & 17.7 & 1.7  & 22.31 & 11.7 \\
                              & w/o Anc                        & 3.1 & 50.5 & 13.4 & 1.1  & 18.74 & 10.9 \\
                              & w/o RD                         & 2.5 & 51.2 & 12.4 & 1.1  & 17.35 & 8.41 \\
                              & w/o RD\&Anc                    & 3.1 & 50.9 & 13.8 & 1.1  & 18.91 & 8.62 \\
\bottomrule
\end{tabular}
\caption{Ablation studies on the preferred response ($y^+$), Reflective-Distillation (RD) loss, and Anchored Regularization (Anc) loss, illustrating their respective contributions to overall model performance.}
\label{tab:ablation_1}
\end{table}

\begin{table}[t]
\centering
\scriptsize
\setlength{\tabcolsep}{1pt}
\renewcommand{\arraystretch}{1.15}
\begin{tabular}{llcccc|cc}
\toprule
\multicolumn{2}{c}{} & \multicolumn{4}{c}{\textbf{AMBER}} & \multicolumn{2}{c}{\textbf{Object Hal}}\\
\cmidrule(lr){3-6}\cmidrule(l){7-8}
\textbf{\makecell{Model\\size}} & \textbf{Source} & \textbf{CHAIR}\,$\downarrow$ & \textbf{Cover}\,$\uparrow$ & \textbf{HalRate}\,$\downarrow$ & \textbf{Cog}\,$\downarrow$ & \textbf{CHAIRs}\,$\downarrow$ & \textbf{CHAIRi}\,$\downarrow$\\
\midrule
\multirow{5}{*}{\textbf{7B}}
 & GPT-4V                           & 2.0 & 49.5 & 10.5 & 0.9  & 12.50 & 3.90 \\
 & N/A                              & 3.8 & 46.4 & 14.5 & 1.3  & 13.79 & 6.55 \\
 & GLM-4V                           & 2.1 & 49.4 & 10.8 & 0.9  & 12.79 & 4.01 \\
 & QwenVL                           & 2.3 & 48.7 & 11.0 & 1.0  & 13.12 & 4.35 \\

\midrule
\multirow{5}{*}{\textbf{13B}}
& GPT-4V                           & 2.2 & 53.3 & 11.9 & 0.8  & 15.50 & 5.49 \\
 & N/A                             & 4.1 & 47.2 & 15.6 & 1.3  & 18.24 & 7.43 \\
 & GLM-4V                          & 2.2 & 53.4 & 12.1 & 0.9  & 16.13 & 6.03 \\
 & QwenVL                          & 2.4 & 52.3 & 13.0 & 1.1  & 16.79 & 6.64 \\
\bottomrule
\end{tabular}
\caption{Ablation studies on different reflection sources, including no reflection (N/A), reflection from GLM-4V, and reflection from Qwen3-VL-8B-Instruct.}
\label{tab:ablation_2}
\end{table}
We emphasize that each component of the proposed RPO framework plays a vital role in achieving effective preference alignment. Specifically, we investigate the contribution of the Reflection-Distillation (RD) loss, the Anchored Regularization (Anc) loss, and the reflection mechanism itself, including the impact of reflection quality. All analyses are conducted on both LLaVA-Instruct-1.5-7B and LLaVA-Instruct-1.5-13B models.

To assess the importance of each objective component, we perform a gradual ablation by removing individual loss terms and also compare with method using ground truth answer as preferred response, as reported in Table~\ref{tab:ablation_2}. The results reveal a consistent degradation in performance when either the RD loss or the Anc loss is excluded, confirming that both terms contribute significantly to the overall optimization process and help stabilize preference learning. The results show a consistent degradation in performance when either the RD loss or the Anc loss is removed, confirming that both terms play essential roles in the optimization process and help stabilize preference learning. These findings highlight that reflective distillation, which guides the model toward the unconditional generation distribution, and anchored regularization, which prevents policy drift during optimization, are both crucial for achieving stable convergence and reliable alignment.

To further highlight the role of the reflection-guided preference pair construction, we examine the influence of the reflection mechanism and the quality of the reflection source (teacher model). We compare multiple teacher models, including GPT-4V, GLM-4V, and Qwen3-VL-8B-Instruct, and present the results in Table~\ref{tab:ablation_2}. As observed, the model trained without reflection performs markedly worse than all reflective variants, underscoring the limitation of the standard DPO pair-construction process. Moreover, the performance varies across reflection sources: while the locally deployed Qwen3-VL-8B-Instruct provides relatively limited gains, it still surpasses the non-reflective baseline. These findings demonstrate the critical role of the reflection mechanism, where higher-quality reflections enable more precise preference guidance and yield stronger alignment with human-preferred responses.




%% file: sec/5_conclusion.tex
\section{Conclusion}
In this work, we identify a fundamental limitation of the widely used Self-Evolution DPO paradigm: its preference pairs exhibit low contrastiveness and weak preference signals, which restrict the effectiveness of preference optimization. We propose \textbf{Reflective Preference Optimization (RPO)}, a new on-policy framework that leverages hint-guided reflection to construct preference pairs with substantially stronger discriminative signals.

We provide theoretical analysis showing that reflection amplifies the expected preference margin and alleviates the optimization difficulties inherent to Self-Evolution DPO. To further stabilize training, we introduce a compound objective that integrates preference alignment, reflection-guided distillation, and anchored regularization.

Extensive experiments on both LLaVA-Instruct-1.5-7B and LLaVA-Instruct-1.5-13B demonstrate that RPO achieves state-of-the-art performance across multiple hallucination benchmarks, while requiring significantly fewer optimization steps. Ablation studies confirm the necessity of each component in the framework.

Overall, RPO establishes a robust and sample-efficient paradigm for preference optimization, offering stronger alignment signals, faster convergence, and consistently superior empirical results.